\documentclass[a4paper, conference]{IEEEtran}
\IEEEoverridecommandlockouts
\usepackage{cite}
\usepackage{amsmath,amssymb,amsfonts}
\usepackage{graphicx}
\usepackage{textcomp}
\usepackage{xcolor}
\def\BibTeX{{\rm B\kern-.05em{\sc i\kern-.025em b}\kern-.08em
    T\kern-.1667em\lower.7ex\hbox{E}\kern-.125emX}}

\usepackage{mathptmx}
\usepackage{algorithm}
\usepackage{algpseudocode}
\usepackage{float}
\usepackage{physics}
\usepackage{tabularx}
\usepackage{array}
\usepackage{multirow}
\usepackage{booktabs}
\ifCLASSOPTIONcompsoc
\usepackage[caption=false,font=normalsize,labelfont=sf,textfont=sf]{subfig}
\else
\usepackage[caption=false,font=footnotesize]{subfig}
\fi
\begin{document}

\title{A sensor fusion approach for improving implementation speed and accuracy of RTAB-Map algorithm based indoor 3D mapping\\
\thanks{Hoang-Anh Phan, Phuc Vinh Nguyen, Thu Hang Thi Khuat, Hieu Dang Van, Dong Huu Quoc Tran, Tung Thanh Bui, Van Nguyen Thi Thanh, and Trinh Chu Duc are with the Faculty of Electronics and Telecommunications, VNU University of Engineering and Technology, Hanoi, Vietnam.{\tt~anhph@vnu.edu.vn, vinhuet01@gmail.com, thuhangkt01@gmail.com, hieu.dv@vnu.edu.vn, dongtran.robotics@gmail.com, tungbt@vnu.edu.vn, vanntt@vnu.edu.vn, trinhcd@vnu.edu.vn}}
\thanks{Bao Lam Dang is with Hanoi University of Science and Technology, Hanoi, Vietnam.\tt~lam.dangbao@hust.edu.vn}}

\author{\IEEEauthorblockN{Hoang-Anh Phan,  Phuc Vinh Nguyen, Thu Hang Thi Khuat, Hieu Dang Van,\\ Dong Huu Quoc Tran, Bao Lam Dang, Tung Thanh Bui, Van Nguyen Thi Thanh and Trinh Chu Duc}}


\maketitle

\begin{abstract}
In recent years, 3D mapping for indoor environments has undergone considerable research and improvement because of its effective applications in various fields, including robotics, autonomous navigation, and virtual reality. Building an accurate 3D map for indoor environment is challenging due to the complex nature of the indoor space, the problem of real-time embedding and positioning errors of the robot system. This study proposes a method to improve the accuracy, speed, and quality of 3D indoor mapping by fusing data from the Inertial Measurement System (IMU) of the Intel Realsense D435i camera, the Ultrasonic-based Indoor Positioning System (IPS), and the encoder of the robot’s wheel using the extended Kalman filter (EKF) algorithm. The merged data is processed using a Real-time Image Based Mapping algorithm (RTAB-Map), with the processing frequency updated in synch with the position frequency of the IPS device. The results suggest that fusing IMU and IPS data significantly improves the accuracy, mapping time, and quality of 3D maps. Our study highlights the proposed method's potential to improve indoor mapping in various fields, indicating that the fusion of multiple data sources can be a valuable tool in creating high-quality 3D indoor maps.
\end{abstract}

\section{Introduction}
\label{sec:intro}
Autonomous robots are becoming increasingly used for a wide range of indoor applications, including autonomous navigation, inspection, and surveillance. One of the critical requirements for these applications is fast and accurate 3D mapping of the indoor environment.  One of the methods to build 3D maps is Visual  Simultaneous Localization And Mapping (Visual SLAM), which refers to SLAM \cite{durrant2006simultaneous} systems that use the camera as the primary input sensor to receive visual information only \cite{taketomi2017visual}. Several methods can be mentioned as Mono-SLAM - Visual Simultaneous Localization And Mapping technology for real-time applications developed \cite{davison2003real}, PTAM (Parallel Tracking and Mapping) \cite{klein2007parallel}, ORB-SLAM (Oriented FAST and Rotated BRIEF - SLAM) \cite{mur2015orb} or ORB-SLAM2 \cite{mur2017orb}. This feature-based SLAM technology is the foundation of modern SLAMs for real-time applications. Visual SLAM still has difficulties in experimenting with indoor environments such as lighting problems affecting camera quality, and bumpy terrain affecting robot accuracy \cite{aqel2016review}. One of the main challenges in Visual SLAM is the accumulation of errors over time \cite{clipp2010parallel}, causing drift in the estimated camera pose and resulting in inaccuracies in the generated map. Drift can be caused by various factors and requires correction methods such as loop closure detection and sensor fusion techniques to maintain a consistent reference frame.  

To address the challenge of drift in Visual SLAM, researchers have proposed several methods, including Loop closure detection \cite{zhang2017loop}, Sensor fusion \cite{garcia2016indoor}, Keyframe-based SLAM \cite{mur2014fast, klein2008improving}, Robust feature detection, and Matching \cite{kadir2013features}. In recent years, Sensor fusion has become a popular method to improve the accuracy of Visual SLAM. Some fusion techniques are the extended Kalman filter (EKF) \cite{ribeiro2004kalman}, the unscented Kalman filter (UKF) \cite{wan2001unscented}, and Particle filters \cite{thrun2002particle}, to estimate the position and orientation of the camera by fusing data from multiple sensors \cite{nutzi2011fusion}. While these methods are widely used in Visual SLAM, they present certain limitations such as reliance on accurate sensor data, linearization of the system in EKF, high computational complexity in particle filters, and increased computation in UKF.  

This study presents a method that utilizes a fusion of data acquired from multiple sensors to serve as an alternative measurement source to the Visual Odometry (VO) \cite{engel2017direct} of the RTAB-Map method. Specifically, we combine data from an ultrasonic-based Indoor Positioning System (IPS) with wheel encoders, and an Inertial Measurement System (IMU) obtained from a D435i camera by implementing the EKF algorithm. The proposed method aims to address two critical issues in the current system: the drift error in camera pose estimation and the noise from separate sensors, thereby enhancing the accuracy, speed, and quality of indoor 3D mapping. The external positioning source has high accuracy due to ultrasonic technology, making it a valuable tool for improving indoor mapping. Additionally, the high sampling rate of IPS, in contrast to the low sampling rate of VO, which has to be estimated from keyframe images, results in improved 3D mapping quality and time. The experimental results indicate that the proposed method effectively mitigates errors in position and trajectory during the robot's mapping process at high speeds, thereby improving the accuracy of the resulting map. Moreover, the proposed method achieves a faster map update frequency and a more optimal time for constructing the 3D map, resulting in a more informative 3D map compared to the base method - is the RTAB-Map method where the odometry known as VO is calculated from the image information and the IMU of the D435i camera device. 
\section{Propose method}
\label{sec:problem}

\begin{figure*}[htbp]
    \centering
    \subfloat[]{\includegraphics[height=2.4in]{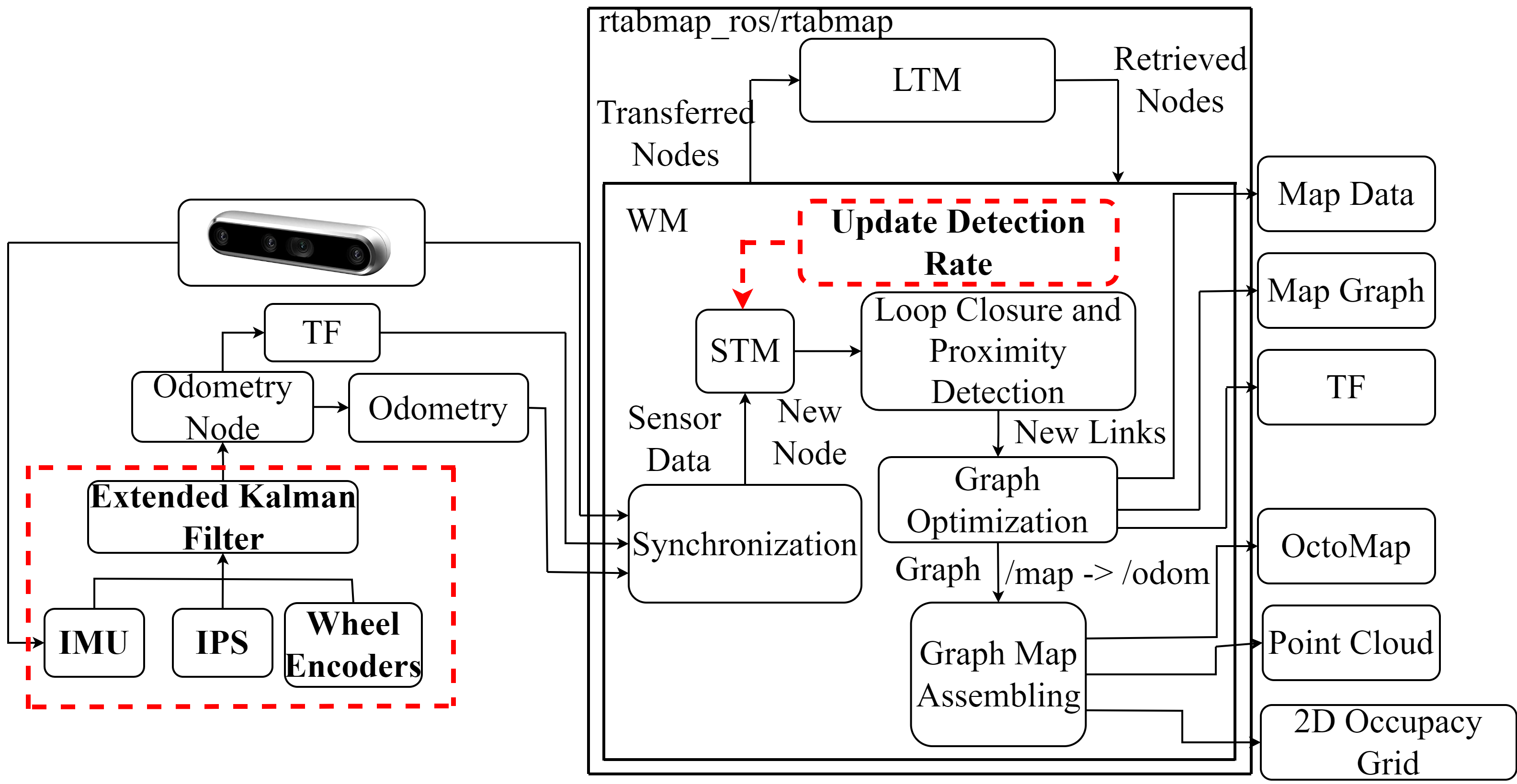}
     \label{fig:rtab_fusion}}
   \centering
    \subfloat[]{\includegraphics[height=2.4in]{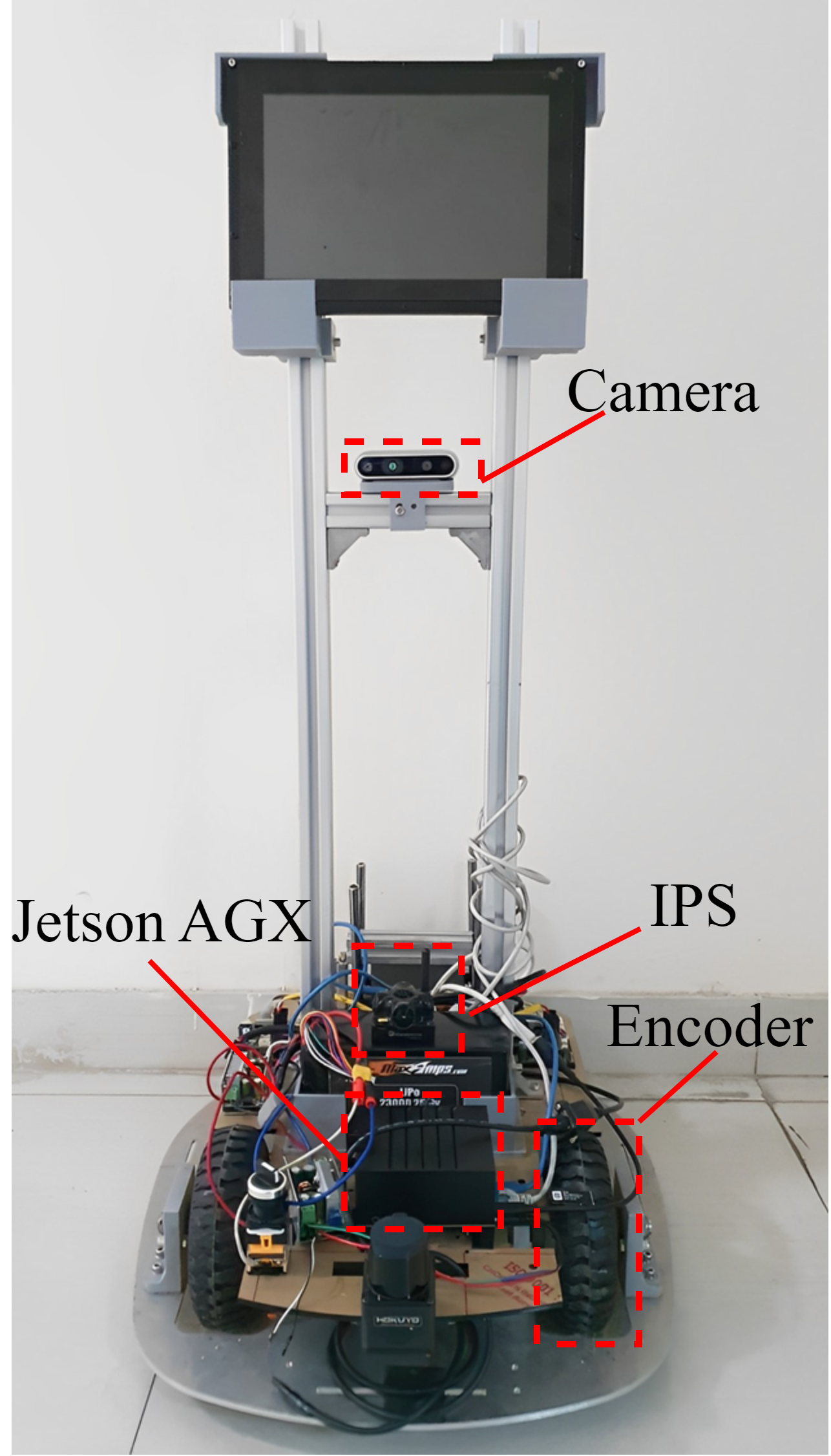}
     \label{fig:robot}}
    \captionsetup{justification=centering}
    \caption{The proposed modified RTAB-map algorithm. (a) Algorithm flowchart (b) A robot platform for implementing and examining the proposed algorithm.}
    \label{fig:map}
\end{figure*}%
In this study, a differential two-wheel indoor mobile robot platform was utilized for the experimental setup (Fig. \ref{fig:robot}), which ensures stability during robot movement. The system was deployed on a Jetson AGX embedded computer, with the input data obtained from the depth camera D435i for Visual-SLAM. The Ultrasonic-based IPS employed for the proposed method consisted of five beacons, including four static beacons fixed to the walls and one dynamic beacon mounted on the robot to connect and transfer position data. Fig. \ref{fig:rtab_fusion} illustrates the flowchart of the proposed RTAB-Map algorithm, which includes the fusion of data from multiple sensor sources such as ``IPS'', ``Wheel Encoders'' and ``IMU'' using an extended Kalman filter. 

The study aimed to provide an alternative source of measurement with higher accuracy and reliability than visual odometry by incorporating an Ultrasonic-based IPS into the system. Moreover, to enhance the quality of 3D maps, the ``Update Detection Rate'' block was added to the main algorithm to update the algorithm with high frequency synchronized with the IPS device. The extended Kalman filter was selected as a commonly used method for sensor fusion in this study. The input format includes coordinates $(x,y)$ from ``IPS'', velocities $(v_x, v_y)$ from ``Wheel Encoders'', orientation $(yaw)$; angular velocities $(v_{yaw})$ and accelerations $(a_x, a_y)$ measured by ``IMU'' of camera D435i. The EKF algorithm consists of two main steps: the prediction step and the update step, the algorithm is implemented as follows:  

•	The prediction step:

The equation that predicts the robot's state at the current time step is based on its previous state estimate and the control input.

\begin{equation}
x_{\{k \mid k-1\}}=f\left(x_{\{k-1\}}, u_k\right)
\end{equation}

\begin{equation}
P_{\{k \mid k-1\}}=F_k * P_{\{k-1 \mid k-1\}} * F_k^T+Q_k
\end{equation}

Where ${x_{\{k-1\}}}$ is the previous state estimate ${x_{\{k \mid k-1\}}}$ is the predicted state estimate, ${u_k}$ is the control input at time step $k$, ${F_k}$ is the Jacobian matrix of the motion model with respect to the state vector at time step $k$, $P_{\{k-1 \mid k-1\}}$ is the covariance matrix of the previous state estimate, $P_{\{k \mid k-1\}}$ is predicted covariance matrix and $Q_k$ is a manually constructed matrix that represents the noise that occurs throughout the estimate process.

•	The update step:

The equation that predicts the sensor's measurements at time step $k$:
\begin{equation}
z_k=h\left(x_{\{k \mid k-1\}}\right)+v_k
\end{equation}
Kalman gain calculation:
\begin{equation}
K_k=P_{\{k \mid k-1\}} * H_k^T *\left(H_k * P_{\{k \mid k-1\}} * H_k^T+R_k\right)^{\{-1\}}
\end{equation}
Then update the state estimate:
\begin{equation}
x_{\{k \mid k\}}=x_{\{k \mid k-1\}}+K_k *\left(z_k-h\left(x_{\{k \mid k-1\}}\right)\right)
\end{equation}
Update the state estimate covariance matrix:
\begin{equation}
P_{\{k \mid k\}}=\left(I-K_k * H_k\right) * P_{\{k \mid k-1\}}
\end{equation}

Where $z_k$ is sensor measurement at time step $k$, $h\left(x_{\{k \mid k-1\}}\right)$ is expected sensor measurement based on the predicted state estimate, $v_k$ is measurement noise, $H_k$ is the Jacobian matrix of the measurement model with respect to the state vector at time step $k$, $R_k$ is measurement noise covariance matrix, $K_k$ is Kalman gain matrix, $x_{\{k \mid k\}}$ is updated state estimate and $P_{\{k \mid k\}}$ is updated covariance matrix.

 Finally, the output provides filtered odometry data then the data is sent along with D435i camera image information to the ``RTAB-Map" algorithm block.

The results show that the proposed method provides a more accurate estimate of the robot's position, which will be presented in Sec \ref{sec:result}.

The EKF generates filtered odometry data as its output, which is subsequently transmitted to the ``RTAB-Map'' algorithm block together with image information from the D435i camera. In the Working Memory (WM) of the RTAB- Map (Fig. \ref{fig:rtab_fusion}), the Short-term Memory (STM) block has the function of storing odometry data and raw sensor data. When a new node is created in STM the local occupancy grid is now calculated from the depth image \cite{labbe2019rtab}. The nodes in the STM will send data at a rate set in the RTAB-Map algorithm. Two parameters affect the updating of data from STM and updating of map data including:  

• ``Rtabmap/MapUpdateDetectionRate'': This parameter relates to how often the map is updated based on new sensor data and represents the maximum rate (in Hz).  

• ``Rtabmap/DetectionRate'': This parameter relates to the rate at which RTAB-Map performs feature detection on incoming D435i camera data and represents the maximum rate (in Hz). 

By default, the odometry input of the ``Loop Closure and Proximity Detection" block only uses the depth camera, resulting in a low update frequency of 1 Hz due to delays in visual odometry estimation. This causes real-time delays in the system, leading to errors in the map and slow updating of point cloud data. To improve the system's performance, a new node in the STM is created that communicates solely with the ``Loop Closure and Proximity Detection" block and receives odometry data fused from the IPS and IMU, increasing the update rate up to 10 Hz. The proposed method adjusted the ``Rtabmap/MapUpdateDetectionRate" and ``Rtabmap/DetectionRate" parameters to match the filtered odometry source's update rate, ensuring accurate and timely map updating. 
\section{Experiments and Results}
\label{sec:result}
\subsection{Estimated Position Error}

\begin{figure}[htbp]
    \centering
    \subfloat[]{\includegraphics[height=1.9in]{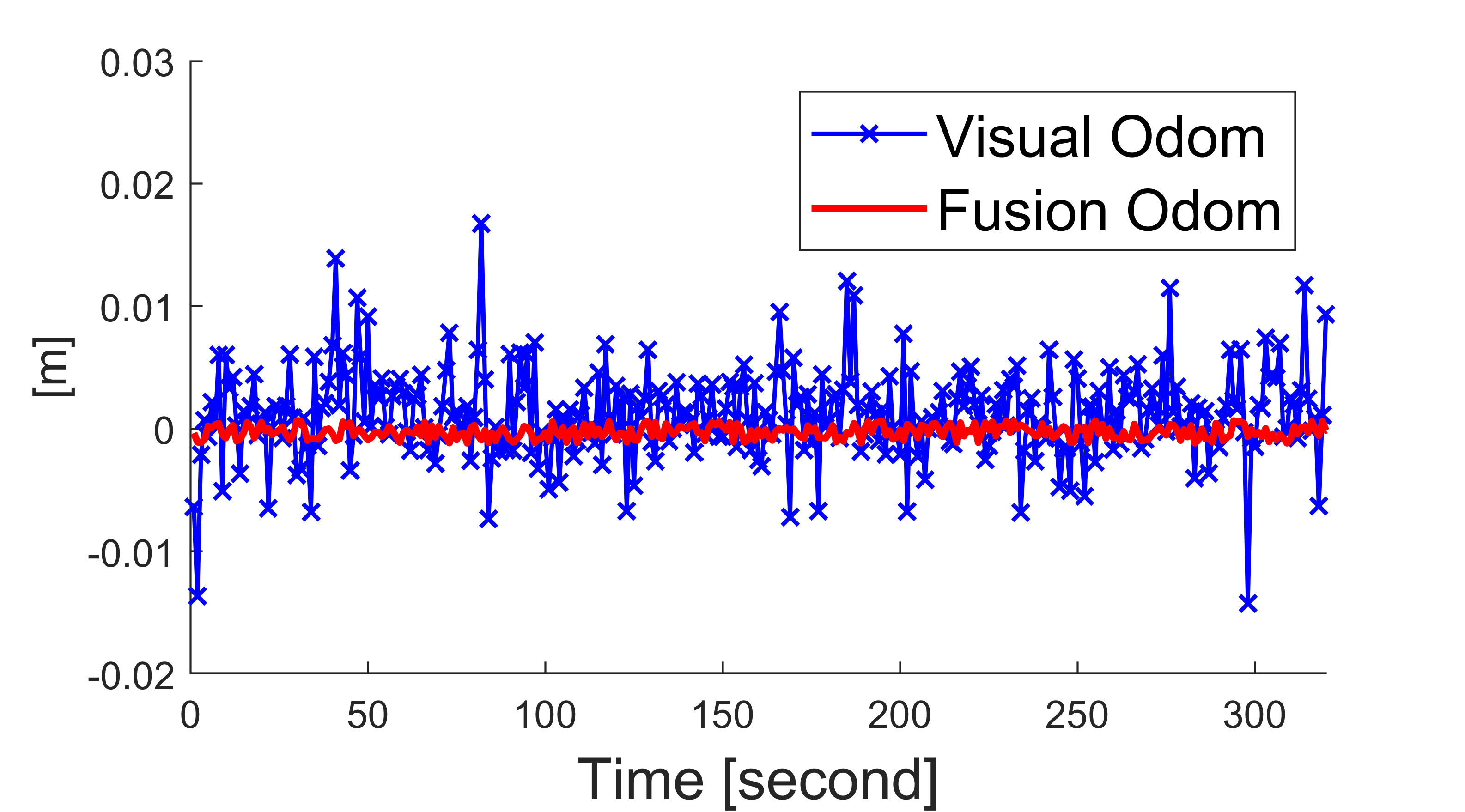}
     \label{fig:Pose_error_X}}\\
   \centering
    \subfloat[]{\includegraphics[height=1.9in]{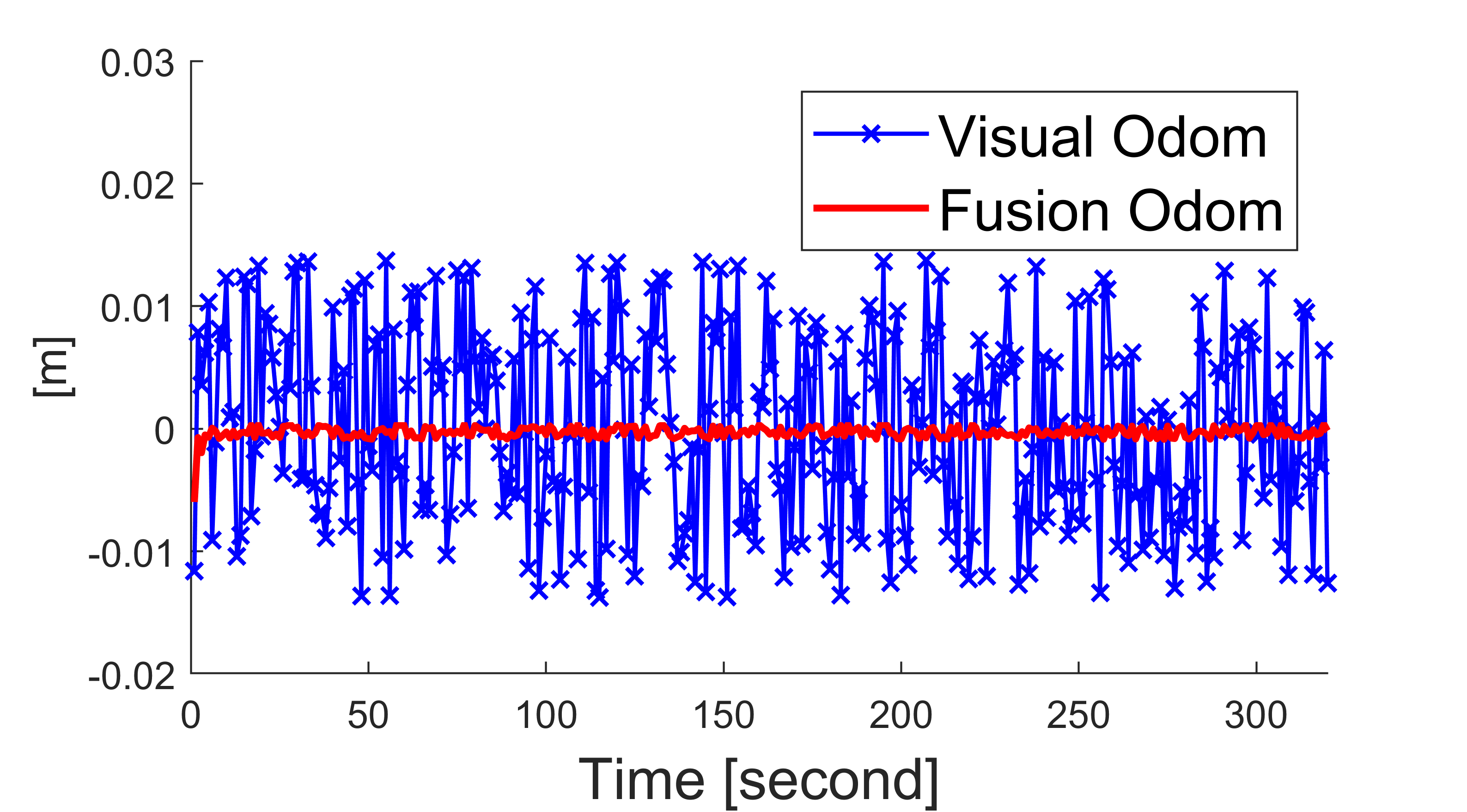}
     \label{fig:Pose_error_Y}}
    \captionsetup{justification=centering}
    \caption{Estimated position error. a) Pose error X-axis b) Pose error Y-axis }
    \label{fig:pose_err}
\end{figure}

In this case, the comparison was performed on the position error of the robot while stationary for 60 seconds between ``Visual Odom'' (Blue line) - representing the base method, and ``Fusion Odom'' (Red line) - representing the proposed  method, as shown in Fig. \ref{fig:Pose_error_X} and  Fig. \ref{fig:Pose_error_Y}, respectively. 

The comparison in the selected 60-second interval can show the drift of ''Visual Odom''. It could be observed from the line graph that the position estimate in the base method exhibits continuous jumps over time during the 3D map construction process. However, after sensor fusion, the position estimate becomes more accurate and stable during a specific time period.
\subsection{Estimated Trajectory Error}

\begin{figure}[htbp]
    \centering
    \subfloat[]{\includegraphics[height=2.1in]{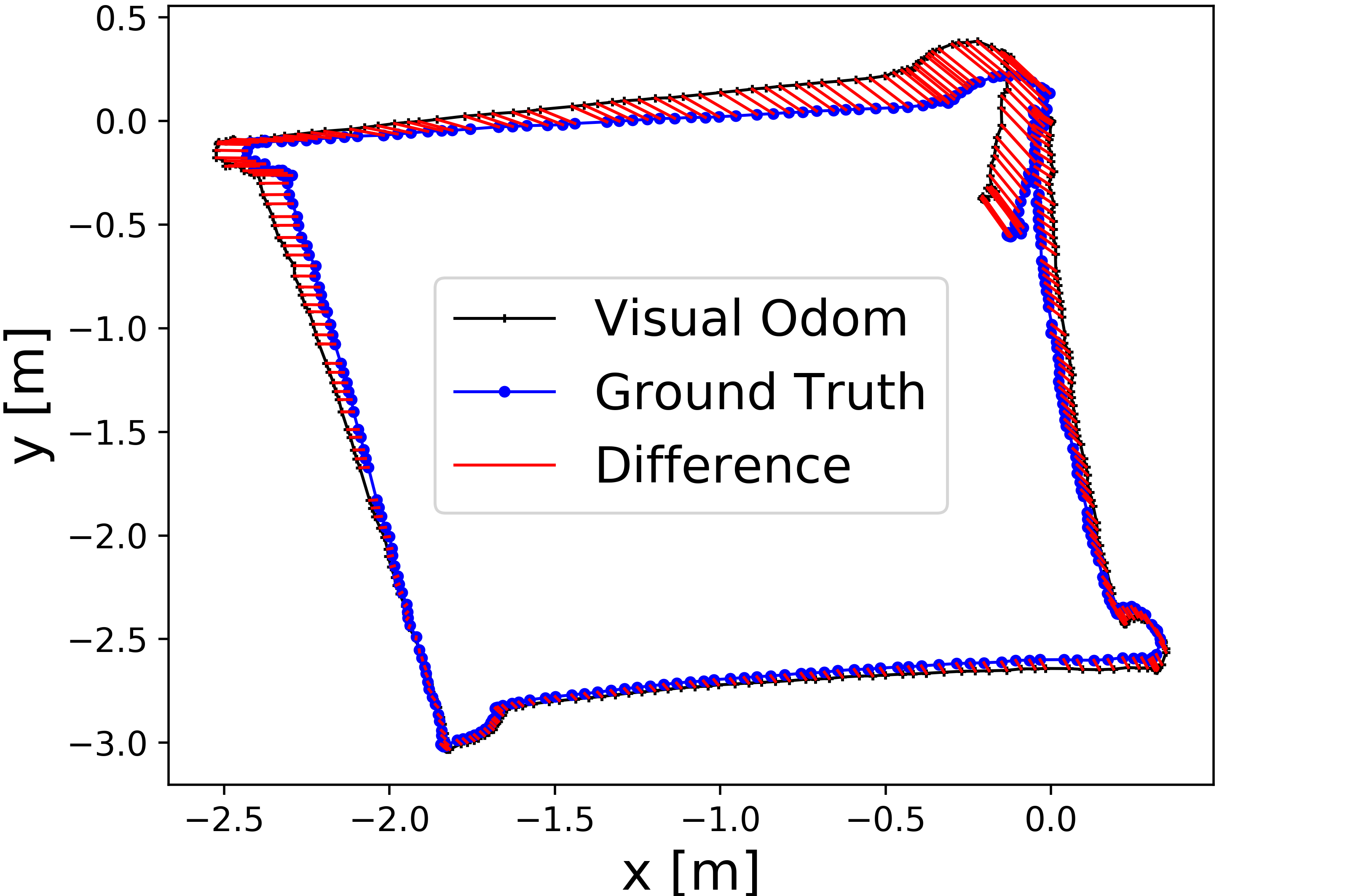}
     \label{fig:noips3}}\\
    \centering
    \subfloat[]{\includegraphics[height=2.1in]{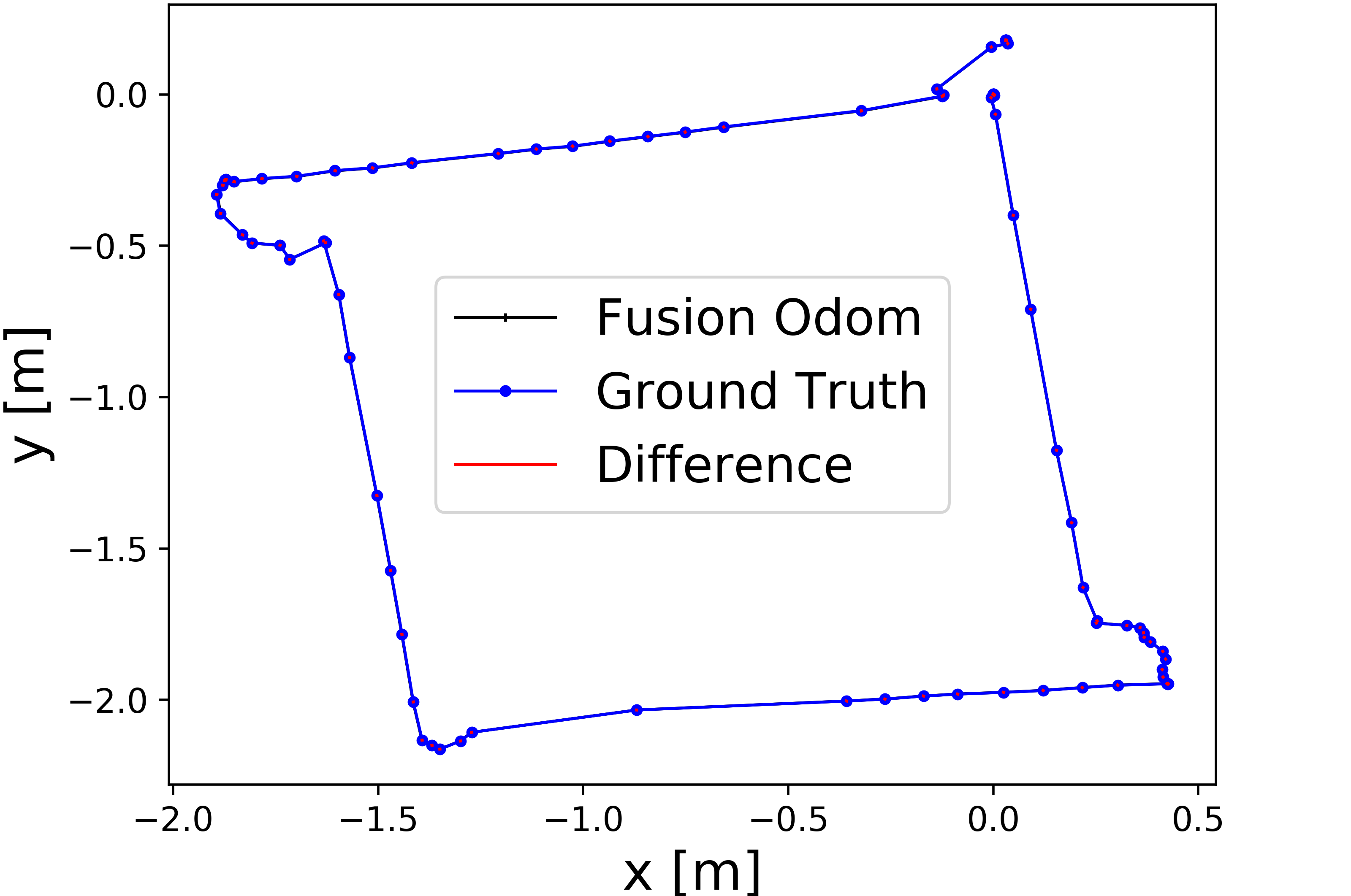}
     \label{fig:ips3}}\\
   \centering
    \subfloat[]{\includegraphics[height=1.7in]{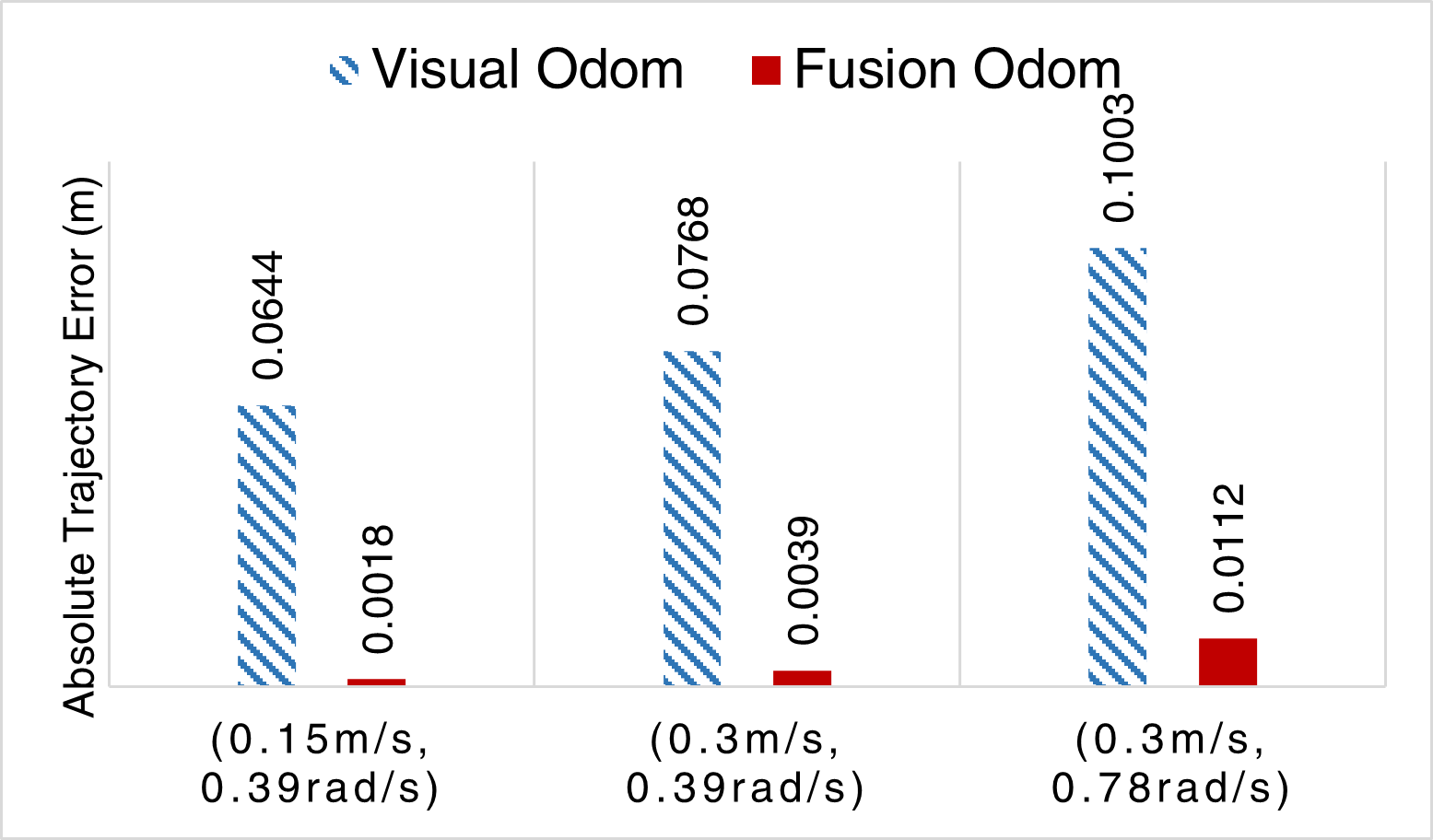}
     \label{fig:ate}}
    \captionsetup{justification=centering}
    \caption{Estimated trajectory error in $v = 0.3 m/s$ and $\omega = 0.78 rad/s$. a) Trajectory error case in the base method. b) Trajectory error case in the proposed  method. c) Estimated absolute trajectory error.}
    \label{fig:Trajec}
\end{figure}

A comparative analysis of the trajectories obtained from both methods was conducted, and the results were highly favorable. The mapping process was carried out at different speeds of the robot with a linear velocity of $v = 0.15 m/s$, $0.3 m/s$ and rotational velocity of $\omega = 0.39 rad/s$, $0.78 rad/s$. The trajectory error was estimated with the ground truth taken from the 3D map after the mapping process was completed. The results of the $v = 0.3 m/s$ and $\omega = 0.78 rad/s$ experiments are shown in Fig. \ref{fig:Trajec}. It was observed that there were significant differences between the trajectory of ``Visual Odom'' and the ground truth (Fig. \ref{fig:noips3}), which was caused by pose drift when the robot turned. However, ``Fusion Odom'' performed outstandingly (Fig. \ref{fig:ips3}) in reducing the trajectory error.

In order to evaluate the accuracy of the proposed method, the Absolute Trajectory Error (ATE) \cite{zhang2018tutorial} was used in this study. The ATE was calculated as the average of the Euclidean distance difference between the ground truth and the supplied odometry input over the total number of time steps.
Formula to calculate ATE:
\begin{equation}
    ATE=\frac{1}{N} \sum_{i=1}^N\left\|P_i-Q_i\right\|
\end{equation}
Where $N$ is the total number of time steps, $\left\|P_i-Q_i\right\|$ is the error of the Euclidean distance of the two methods.

The results were shown in Fig. \ref{fig:ate}, where the ``Fusion Odom'' column represented the ATE coefficient of the proposed method, and the ``Visual Odom'' column represented the base method. The ATE coefficient of the proposed method was significantly lower than that of the base method, ranging from 9-36 times lower. These findings suggested that the absolute orbital error of the proposed method had been greatly reduced compared to the base method. Additionally, as the linear and angular velocities increased, the orbital error increased significantly. At $(v=0.3 m/s$, $\omega=0.78 rad/s)$, the error of the base method was approximately $0.1$m, whereas the error of the proposed method was only approximately $0.01 m$. These results demonstrate the efficacy of the proposed method in reducing errors and improving accuracy.

\subsection{3D map quality and processing time}

\begin{figure}[htbp]
    \centering
    \subfloat[]{\includegraphics[height=1.45in]{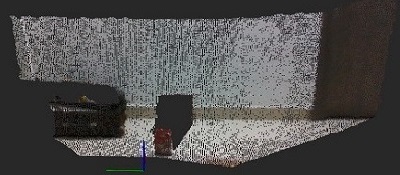}
     \label{fig:qua_bas}}\\
   \centering
    \subfloat[]{\includegraphics[height=1.45in]{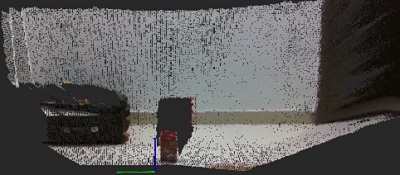}
     \label{fig:qua_pro}}
    \captionsetup{justification=centering}
    \caption{3D mapping quality a) Base method scans 83902 cloud points. b) Proposed  method scans 188190 cloud points. }
    \label{fig:qua_map}
\end{figure}

A comparative study (Fig. \ref{fig:qua_map}) was conducted to investigate the effectiveness of synchronizing the frequency of updating the map processing of the RTAB-Map method with that of location updates from the IPS. The experiment involved scanning an area of the environment with a robot over a fixed time interval ($\delta t = 2s$) and a fixed angular velocity ($\omega = 0.2rad/s$). The number of point clouds on the 3D map generated using the base method and the proposed method was compared. The results showed that the proposed method generated a higher number of point clouds (188190 points) in Fig. \ref{fig:qua_pro} compared to the base method (83902 points) in Fig. \ref{fig:qua_bas}, indicating an improvement in the quality of the 3D map. The findings demonstrated the clear benefits of increasing the frequency of updating the map processing of RTAB-Map to be synchronized with the frequency of location updates from the ``Fusion Odom''.

\begin{figure}[htbp]
    \centering
\includegraphics[height=1.9in]{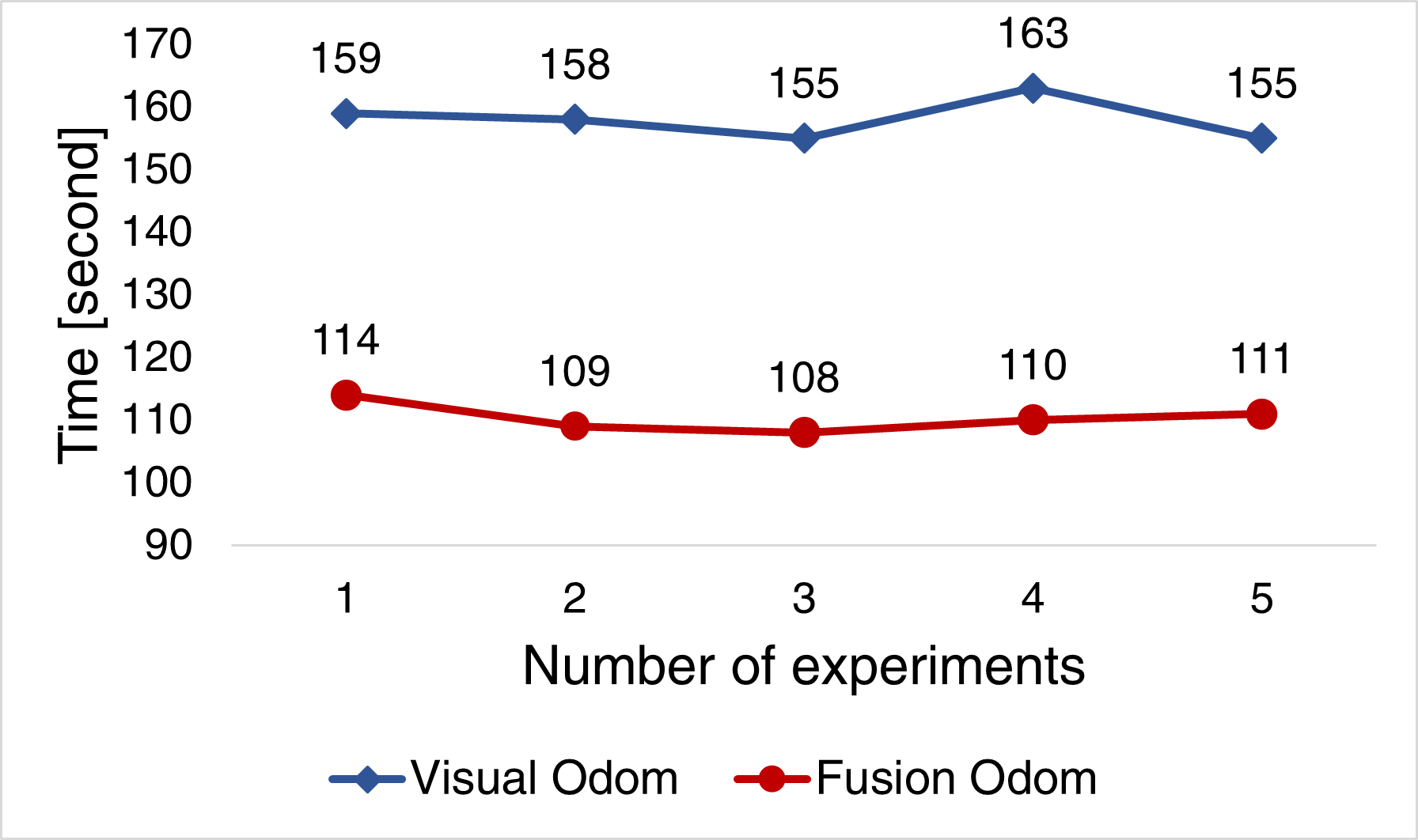}
\caption{Time of 3D map building}
    \label{fig:time_build_map}
\end{figure}

During testing for the robot's construction of 3D maps using two different methods on the same trajectory with identical linear and angular velocities, the interference and loss of odometry in the control process led to the repositioning of the robot by the operator and adjustment of the map construction trajectory by the RTAB-Map system. This interference significantly affected the time required to build the map. A test was conducted to assess the time taken by the robot to build a 3D map with the trajectory obtained after 5 test runs, present in Fig. \ref{fig:time_build_map}. The results indicated that the proposed method had an average mapping time of $110.4$ seconds, while the base method took an average of $158$ seconds to execute.
\section{Conclusion}
In this study, we proposed an approach to enhance the accuracy, speed, and quality of indoor 3D mapping by integrating data from an external ultrasonic-based Indoor Positioning System (IPS), Inertial Measurement System (IMU) from a D435i camera, and wheel encoders. The integration is achieved using the extended Kalman filter, which serves as an effective means to fuse the data obtained from these sensors. The proposed method aims to reduce the drift error of the camera pose and diminish noise from separate sensors. The study evaluates and compares the base method and the proposed method based on criteria such as position and trajectory estimation errors, map accuracy, robot mapping time, and the amount of information in the 3D map. The results indicate that the proposed method reduces errors in position and trajectory estimation while mapping at high speeds, leading to improved map accuracy. The proposed method also optimizes the time required to build the 3D map, increases the map update frequency, and enhances the amount of information contained in the 3D map compared to the base method. The obtained results show the potential of using the proposed method to enhance indoor 3D mapping in robotics, signifying that the fusion of multiple data sources can serve as a valuable tool in generating accurate and high-quality 3D maps.

\bibliographystyle{IEEEtran}
\bibliography{IEEEabrv, ref}

\begin{thebibliography}{10}
\providecommand{\url}[1]{#1}
\csname url@samestyle\endcsname
\providecommand{\newblock}{\relax}
\providecommand{\bibinfo}[2]{#2}
\providecommand{\BIBentrySTDinterwordspacing}{\spaceskip=0pt\relax}
\providecommand{\BIBentryALTinterwordstretchfactor}{4}
\providecommand{\BIBentryALTinterwordspacing}{\spaceskip=\fontdimen2\font plus
\BIBentryALTinterwordstretchfactor\fontdimen3\font minus
  \fontdimen4\font\relax}
\providecommand{\BIBforeignlanguage}[2]{{%
\expandafter\ifx\csname l@#1\endcsname\relax
\typeout{** WARNING: IEEEtran.bst: No hyphenation pattern has been}%
\typeout{** loaded for the language `#1'. Using the pattern for}%
\typeout{** the default language instead.}%
\else
\language=\csname l@#1\endcsname
\fi
#2}}
\providecommand{\BIBdecl}{\relax}
\BIBdecl

\bibitem{durrant2006simultaneous}
H.~Durrant-Whyte and T.~Bailey, ``Simultaneous localization and mapping: part
  i,'' \emph{{IEEE} Robot. Autom. Mag.}, vol.~13, no.~2, pp. 99--110, 2006.

\bibitem{taketomi2017visual}
T.~Taketomi, H.~Uchiyama, and S.~Ikeda, ``Visual slam algorithms: A survey from
  2010 to 2016,'' \emph{IPSJ Transactions on Computer Vision and Applications},
  vol.~9, no.~1, pp. 1--11, 2017.

\bibitem{davison2003real}
A.~J. Davison, ``Real-time simultaneous localisation and mapping with a single
  camera,'' in \emph{Computer Vision, IEEE International Conference on},
  vol.~3.\hskip 1em plus 0.5em minus 0.4em\relax IEEE Computer Society, 2003,
  pp. 1403--1403.

\bibitem{klein2007parallel}
G.~Klein and D.~Murray, ``Parallel tracking and mapping for small ar
  workspaces,'' in \emph{2007 6th IEEE and ACM international symposium on mixed
  and augmented reality}.\hskip 1em plus 0.5em minus 0.4em\relax IEEE, 2007,
  pp. 225--234.

\bibitem{mur2015orb}
R.~Mur-Artal, J.~M.~M. Montiel, and J.~D. Tardos, ``Orb-slam: a versatile and
  accurate monocular slam system,'' \emph{{IEEE} Trans. Robot.}, vol.~31,
  no.~5, pp. 1147--1163, 2015.

\bibitem{mur2017orb}
R.~Mur-Artal and J.~D. Tard{\'o}s, ``Orb-slam2: An open-source slam system for
  monocular, stereo, and rgb-d cameras,'' \emph{IEEE transactions on robotics},
  vol.~33, no.~5, pp. 1255--1262, 2017.

\bibitem{aqel2016review}
M.~O. Aqel, M.~H. Marhaban, M.~I. Saripan, and N.~B. Ismail, ``Review of visual
  odometry: types, approaches, challenges, and applications,''
  \emph{SpringerPlus}, vol.~5, pp. 1--26, 2016.

\bibitem{clipp2010parallel}
B.~Clipp, J.~Lim, J.-M. Frahm, and M.~Pollefeys, ``Parallel, real-time visual
  slam,'' in \emph{2010 IEEE/RSJ International Conference on Intelligent Robots
  and Systems}.\hskip 1em plus 0.5em minus 0.4em\relax IEEE, 2010, pp.
  3961--3968.

\bibitem{zhang2017loop}
X.~Zhang, Y.~Su, and X.~Zhu, ``Loop closure detection for visual slam systems
  using convolutional neural network,'' in \emph{2017 23rd International
  Conference on Automation and Computing (ICAC)}.\hskip 1em plus 0.5em minus
  0.4em\relax IEEE, 2017, pp. 1--6.

\bibitem{garcia2016indoor}
S.~Garc{\'\i}a, M.~E. L{\'o}pez, R.~Barea, L.~M. Bergasa, A.~G{\'o}mez, and
  E.~J. Molinos, ``Indoor slam for micro aerial vehicles control using
  monocular camera and sensor fusion,'' in \emph{2016 international conference
  on autonomous robot systems and competitions (ICARSC)}.\hskip 1em plus 0.5em
  minus 0.4em\relax IEEE, 2016, pp. 205--210.

\bibitem{mur2014fast}
R.~Mur-Artal and J.~D. Tard{\'o}s, ``Fast relocalisation and loop closing in
  keyframe-based slam,'' in \emph{2014 IEEE International Conference on
  Robotics and Automation (ICRA)}.\hskip 1em plus 0.5em minus 0.4em\relax IEEE,
  2014, pp. 846--853.

\bibitem{klein2008improving}
G.~Klein and D.~Murray, ``Improving the agility of keyframe-based slam,'' in
  \emph{Computer Vision--ECCV 2008: 10th European Conference on Computer
  Vision, Marseille, France, October 12-18, 2008, Proceedings, Part II
  10}.\hskip 1em plus 0.5em minus 0.4em\relax Springer, 2008, pp. 802--815.

\bibitem{kadir2013features}
H.~A. Kadir and M.~R. Arshad, ``Features detection and matching for visual
  simultaneous localization and mapping (vslam),'' in \emph{2013 IEEE
  International Conference on Control System, Computing and Engineering}.\hskip
  1em plus 0.5em minus 0.4em\relax IEEE, 2013, pp. 40--45.

\bibitem{ribeiro2004kalman}
M.~I. Ribeiro, ``Kalman and extended kalman filters: Concept, derivation and
  properties,'' \emph{Institute for Systems and Robotics}, vol.~43, no.~46, pp.
  3736--3741, 2004.

\bibitem{wan2001unscented}
E.~A. Wan and R.~Van Der~Merwe, ``The unscented kalman filter,'' \emph{Kalman
  filtering and neural networks}, pp. 221--280, 2001.

\bibitem{thrun2002particle}
S.~Thrun, ``Particle filters in robotics.'' in \emph{UAI}, vol.~2.\hskip 1em
  plus 0.5em minus 0.4em\relax Citeseer, 2002, pp. 511--518.

\bibitem{nutzi2011fusion}
G.~N{\"u}tzi, S.~Weiss, D.~Scaramuzza, and R.~Siegwart, ``Fusion of imu and
  vision for absolute scale estimation in monocular slam,'' \emph{Journal of
  intelligent \& robotic systems}, vol.~61, no. 1-4, pp. 287--299, 2011.

\bibitem{engel2017direct}
J.~Engel, V.~Koltun, and D.~Cremers, ``Direct sparse odometry,'' \emph{{IEEE}
  Trans. Pattern Anal. Mach. Intell.}, vol.~40, no.~3, pp. 611--625, 2017.

\bibitem{labbe2019rtab}
M.~Labb{\'e} and F.~Michaud, ``Rtab-map as an open-source lidar and visual
  simultaneous localization and mapping library for large-scale and long-term
  online operation,'' \emph{Journal of Field Robotics}, vol.~36, no.~2, pp.
  416--446, 2019.

\bibitem{zhang2018tutorial}
Z.~Zhang and D.~Scaramuzza, ``A tutorial on quantitative trajectory evaluation
  for visual (-inertial) odometry,'' in \emph{2018 IEEE/RSJ International
  Conference on Intelligent Robots and Systems (IROS)}.\hskip 1em plus 0.5em
  minus 0.4em\relax IEEE, 2018, pp. 7244--7251.

\end{thebibliography}
\end{document}